\documentclass[runningheads]{llncs}
\usepackage[T1]{fontenc}
\usepackage{graphicx}
\usepackage[hidelinks, bookmarks=false]{hyperref}
\usepackage{color}

\usepackage{booktabs}
\usepackage{enumitem}
\usepackage{multirow}
\usepackage{marvosym}
\usepackage{amsmath}
\usepackage{amssymb}
\usepackage{microtype}

\begin{document}
\title{CARE: Confidence-Aware Reasoning for Reliable Medical VQA}
%
%


\author{Yuetian Du\inst{1}\textsuperscript{\ensuremath{\dagger}}  
\and
Yucheng Wang\inst{1}\textsuperscript{\ensuremath{\dagger}}  
\and
Zhenyuan Chen\inst{1}  
\and
Luyuan Chen\inst{1}  
\and
Rongyu Zhang\inst{1}  
\and
Jinjian Zhang\inst{2}  
\and
Wei Zhou\inst{2}  
\and
Zhijie Xu\inst{3}  
\and
Ming Kong\inst{1}  
\and
Zhan Zhou\inst{1}  
\and \\
Jie Liu\inst{4}\textsuperscript{\Letter}  
\and
Qiang Zhu\inst{1}\textsuperscript{\Letter}}  

\authorrunning{Y. Du et al.}

\institute{%
\begin{tabular}{@{}c@{}}
\textsuperscript{1} Zhejiang University \quad
\textsuperscript{2} Ant Group \\
\textsuperscript{3} University of Michigan \quad
\textsuperscript{4} City University of Hong Kong \\
\email{\{22421227,zhuq\}@zju.edu.cn} \\
\end{tabular}
}

\maketitle              

\begingroup 
\renewcommand{\thefootnote}{} 
\renewcommand{\theHfootnote}{authornote} \footnotetext{\textsuperscript{\ensuremath{\dagger}} Equal contribution. \quad \textsuperscript{\Letter} Corresponding author.} 
\endgroup 
\setcounter{footnote}{0}
\begin{abstract}
Reinforcement Fine-Tuning (RFT) has enabled medical Multimodal Large Language Models (MLLMs) to produce Chain-of-Thought (CoT) reasoning for visual question answering, yet these models suffer from \textit{confidence miscalibration}---a systematic gap between expressed certainty and actual diagnostic accuracy that undermines clinical trust. We propose \textbf{CARE}, a \textbf{C}onfidence-\textbf{A}ware medical \textbf{RE}asoning framework that jointly optimizes accuracy and calibration through a dual-stage pipeline. First, a scalable Medical-CoT synthesis provides structured cold-start data for Supervised Fine-Tuning. Second, Group Relative Policy Optimization (GRPO) with a novel \textbf{Confidence-Aware Reward (CAR)} mechanism ties the model's confidence to diagnostic correctness within the reward signal. Across three Medical VQA benchmarks, \textbf{CARE} achieves the highest diagnostic accuracy while obtaining the lowest Expected Calibration Error and Hallucination Rate, establishing a foundation for trustworthy clinical decision support. Our code is available at \url{https://github.com/anotherbricki/CARE}.

\keywords{Reinforcement Fine-Tuning \and Medical VQA \and Confidence Calibration.}

\end{abstract}
\section{Introduction}

\begin{figure}[t]
\includegraphics[width=\textwidth]{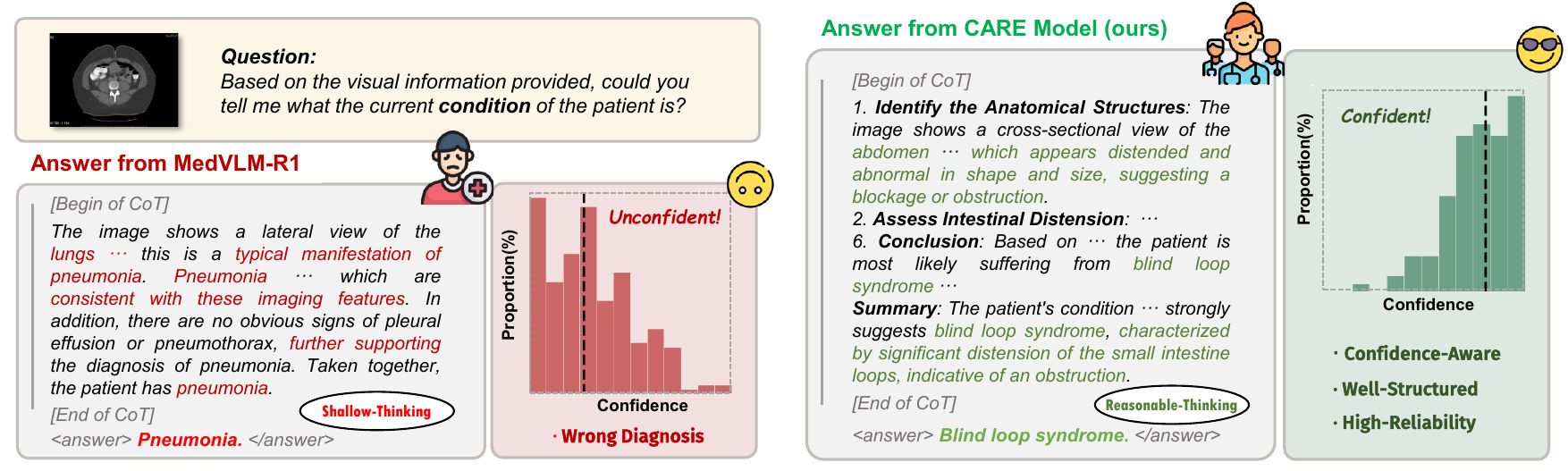}
\caption{\textit{Comparative Analysis of Diagnostic Reasoning Paths.} Existing medical reasoning MLLMs such as MedVLM-R1 exhibit confidence miscalibration, where expressed certainty fails to reflect actual diagnostic accuracy. \textbf{CARE} explicitly aligns subjective confidence with diagnostic correctness, producing verifiable and interpretable reasoning trajectories for clinical decision-making.} \label{fig1}
\end{figure}

Medical Multimodal Large Language Models (MLLMs) are increasingly applied to clinical decision support tasks such as medical visual question answering (VQA)~\cite{li2024llavamed}. Typically built through Supervised Fine-Tuning (SFT) on curated clinical instruction datasets, these models learn a direct input-output mapping that produces predictions without transparent reasoning processes. In clinical practice, where diagnostic errors carry severe consequences, physicians are unlikely to trust model predictions that lack interpretable reasoning chains. This highlights an urgent need for medical MLLMs that can perform transparent, step-by-step clinical reasoning.

Reinforcement Fine-Tuning (RFT), where RL algorithms such as Group Relative Policy Optimization (GRPO)~\cite{shao2024deepseekmathpushinglimitsmathematical} optimize models against verifiable reward signals, has recently enabled reasoning-focused models that generate extended Chain-of-Thought (CoT) outputs. This paradigm has been adopted in the medical domain, producing early medical reasoning models~\cite{lai2025medr1reinforcementlearninggeneralizable,pan2025} for Medical VQA. While these models improve reasoning transparency, a critical issue persists: \textit{confidence miscalibration}~\cite{xiong2023uncertainty,tian2023calibration,du2025confidence}. As illustrated in Figure~\ref{fig1}, existing models exhibit a systematic misalignment between expressed confidence and actual diagnostic accuracy: they may appear underconfident on correct diagnoses or assign unwarranted certainty to incorrect ones. In clinical deployment, such miscalibration undermines trust, as a model that cannot reliably reflect its own uncertainty provides no safe basis for decision-making.

\noindent \textbf{Related Work.} Reinforcement Fine-Tuning (RFT) has emerged as a new paradigm for improving reasoning in large models~\cite{li202512surveyreasoning,xu2025largereasoningmodelssurvey}, surpassing traditional SFT-based CoT approaches. Unlike SFT, which passively fits existing data distributions, RFT enables models to receive direct feedback from verifiable outcomes and dynamically refine their reasoning paths~\cite{chu2025sftmemorizesrlgeneralizes}. Pioneered in mathematical reasoning~\cite{shao2024deepseekmathpushinglimitsmathematical} and popularized by DeepSeek-R1~\cite{deepseekai2025deepseekr1incentivizingreasoningcapability}, RFT has recently been extended to multimodal medical applications, producing models such as Med-R1~\cite{lai2025medr1reinforcementlearninggeneralizable}, MedVLM-R1~\cite{pan2025}, and others~\cite{sun2025enhancingstepbystepverifiablemedical,deria2026medmogroundingunderstandingmultimodal,lasateam2025lingshugeneralistfoundationmodel,huang2025medvlthinkersimplebaselinesmultimodal} for clinical VQA. However, existing RFT frameworks optimize primarily for answer correctness through verifiable rewards~\cite{lambert2025tulu3pushingfrontiers}, without explicitly addressing the alignment between model confidence and prediction accuracy. Unlike methods that elicit verbal confidence or simply reward high confidence, \textbf{CARE} uses confidence as a correctness-conditioned calibration signal within GRPO. Medical-CoT synthesis mainly provides a verified cold start that stabilizes reasoning format and answer extraction for subsequent confidence-aware RL.

To address this gap, we propose \textbf{CARE}---a \textbf{C}onfidence-\textbf{A}ware medical \textbf{RE}asoning framework that jointly optimizes diagnostic accuracy and confidence calibration within a unified two-stage RFT pipeline. Our main contributions are as follows:

\begin{itemize}[label=\textbullet,leftmargin=*]
    \item \textbf{Confidence-Aware Reinforcement Fine-Tuning:} We integrate a novel Confidence-Aware Reward (\textbf{CAR}) into the GRPO framework, explicitly aligning the model's expressed confidence with diagnostic accuracy during RL optimization.

    \item \textbf{Scalable Medical-CoT Data Construction:} We design an automated synthesis pipeline to construct high-quality medical reasoning data for multiple clinical scenarios, providing structured diagnostic trajectories with verifiable conclusions.

    \item \textbf{Consistent Improvements in Accuracy and Calibration:} Evaluations across multiple Medical VQA benchmarks demonstrate that \textbf{CARE} achieves superior diagnostic accuracy while substantially reducing confidence miscalibration, establishing a reliable foundation for clinical decision support.
\end{itemize}

\section{Method}

\subsection{Overview}
\begin{figure}[t]
\includegraphics[width=\textwidth]{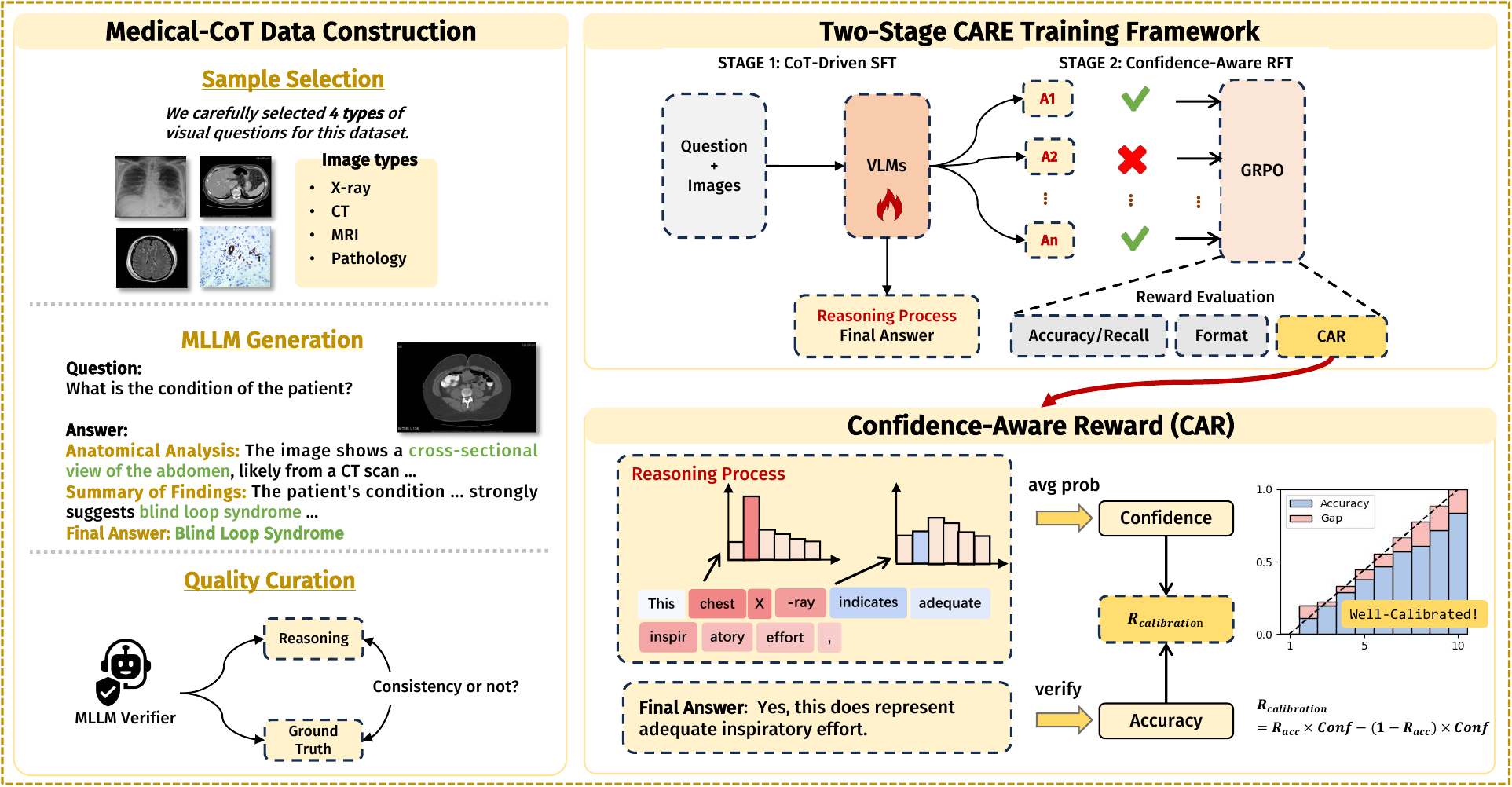}
\caption{\textit{Overview of the \textbf{CARE} Framework.} The pipeline consists of two phases: \textit{(1) Medical-CoT Data Construction} for synthesizing structured diagnostic reasoning paths, and \textit{(2) Two-Stage Optimization}, comprising SFT cold-start and GRPO with the proposed \textbf{Confidence-Aware Reward (CAR)} mechanism to align confidence with diagnostic accuracy.} \label{fig2}
\end{figure}

As illustrated in Figure~\ref{fig2}, the \textbf{CARE} framework consists of two phases: (1) \textit{Medical-CoT Data Synthesis}, which constructs scalable, well-structured diagnostic reasoning paths from existing clinical datasets; and (2) \textit{Two-Stage Optimization}, transitioning from SFT-based domain adaptation to GRPO reinforcement learning driven by a novel \textbf{Confidence-Aware Reward (CAR)} mechanism. This design jointly optimizes diagnostic accuracy and confidence calibration, mitigating the risk of confidence miscalibration in clinical judgments.

\subsection{Scalable Medical-CoT Data Synthesis}
\label{sec:syn}
To facilitate structured reasoning without relying solely on prohibitive expert annotations, we design an automated, reverse-thinking synthesis pipeline. Let $\mathcal{D}_{\text{VQA}} = \{(V_i, Q_i, Y_i)\}_{i=1}^N$ denote a standard Medical-VQA dataset comprising visual contexts $V$, text queries $Q$, and ground truth diagnoses $Y$. We employ a capable base MLLM, denoted as $\pi_{\theta}$, to retroactively generate intermediate reasoning trajectories $\mathcal{T}$:
\begin{equation}
    \mathcal{T}_i \sim \pi_{\theta}(V_i, Q_i, Y_i).
\end{equation}

To ensure the clinical validity and logical coherence of $\mathcal{T}_i$, we enforce a strict structural template requiring explicit phase identifiers (e.g., visual analysis, differential diagnosis) and a conclusive summary. Crucially, rather than relying on raw generative outputs, we implement a rigorous verification mechanism. An auxiliary verifier (e.g., GPT-4o) evaluates each trajectory against the ground truth $Y_i$. A generated trajectory is admitted into the final training corpus $\mathcal{D}_{\text{CoT}}$ if and only if its reasoning chain logically deduces a conclusion that perfectly aligns with $Y_i$. This objective filtering acts as a scalable proxy for quality assurance, ensuring the model learns from goal-oriented, logically sound diagnostic paths.

\subsection{Two-Stage Optimization of CARE}
Leveraging the curated $\mathcal{D}_{\text{CoT}}$, \textbf{CARE} is trained in two stages: SFT cold-start for domain adaptation, followed by confidence-aware GRPO-based reinforcement learning.

\noindent \textbf{Phase I: SFT Cold Start.}
We conduct standard supervised fine-tuning on $\mathcal{D}_{\text{CoT}}$ to establish the model's capacity for structured clinical reasoning. For each tuple $(V, Q, \mathcal{T}, Y) \in \mathcal{D}_{\text{CoT}}$, the model maximizes the likelihood of the target sequence $S = [\mathcal{T}; Y]$:
\begin{equation}
    \mathcal{L}_{\text{SFT}}(\theta) = - \mathbb{E}_{(V,Q,S) \sim \mathcal{D}_{\text{CoT}}} \left[ \sum_{t=1}^{|S|} \log \pi_{\theta}(s_t | V, Q, s_{<t}) \right].
\end{equation}
This phase yields a reference policy $\pi_{\text{ref}}$ that follows the structured reasoning format, serving as initialization for the RL phase.

\noindent \textbf{Phase II: GRPO-based RL.}
We then optimize $\pi_{\theta}$ using GRPO, which eliminates the need for a separate value network. For each query $(V, Q)$, the policy samples $K$ candidate outputs $\{o_1, \dots, o_K\}$. The objective maximizes group-relative advantages while constraining divergence from $\pi_{\text{ref}}$:
\begin{equation}
    J_{\text{GRPO}}(\theta) = \mathbb{E} \left[ \frac{1}{K} \sum_{i=1}^{K} \min \left( \rho_i , \text{clip}(\rho_i, 1-\epsilon, 1+\epsilon)  \right)A_i - \beta \mathbb{D}_{\text{KL}}(\pi_{\theta} \| \pi_{\text{ref}}) \right],
\end{equation}
where $\rho_i = \frac{\pi_{\theta}(o_i | V, Q)}{\pi_{\theta}^{\text{old}}(o_i | V, Q)}$ is the importance weight, $\epsilon$ is the clip ratio, and $\beta$ controls the KL penalty. The advantage $A_i$ is computed by normalizing the composite reward $R_i$ within the sampled group:
\begin{equation}
    A_i = \frac{R_i - \mu(R_{1:K})}{\sigma(R_{1:K})}.
\end{equation}

\subsection{Confidence-Aware Reward (CAR) Formulation}
Existing RFT frameworks rely on binary format or accuracy rewards, leaving confidence miscalibration unaddressed. We introduce the \textbf{Confidence-Aware Reward (CAR)} to close this gap by directly incorporating confidence alignment into the reward signal. The composite reward $R_i = R_{\text{form}} + R_{\text{out}} + R_{\text{calib}}$ consists of three components:

\noindent \textbf{Format \& Output Rewards ($R_{\text{form}}, R_{\text{out}}$).}
$R_{\text{form}} \in \{0, 1\}$ enforces the use of designated \texttt{<think>} and \texttt{<answer>} delimiters. $R_{\text{out}}$ evaluates diagnostic correctness: for closed-ended tasks, it is an exact-match indicator $I(Y \subseteq a_i)$; for open-ended tasks, it uses a recall-based metric to assess the coverage of ground truth $Y$ within the predicted answer $a_i$.

\noindent \textbf{Calibration Reward ($R_{\text{calib}}$).}
This is the core component that distinguishes \textbf{CARE} from standard RFT. We use the model's answer-level predictive confidence as a calibration signal, rather than treating token probability as a complete epistemic uncertainty estimate. Specifically, for each output $o_i$, let $a_i=\{t_1,\ldots,t_{|a_i|}\}$ denote only the tokens inside the \texttt{<answer>} span, excluding reasoning tokens, formatting tokens, and special tokens. We compute:
\begin{equation}
    C(a_i) = \frac{1}{|a_i|} \sum_{j=1}^{|a_i|} \pi_{\theta}(t_j \mid V, Q, t_{<j}).
\end{equation}
The calibration reward ties this answer-level confidence to diagnostic correctness:
\begin{equation}
    R_{\text{calib}}(o_i, Y) = R_{\text{out}} \cdot C(a_i) - \lambda (1 - R_{\text{out}}) \cdot C(a_i),
\end{equation}
where $\lambda$ controls the penalty on overconfident incorrect predictions. When the prediction is correct, higher answer confidence is rewarded; when it is incorrect, high confidence is penalized. This correctness-conditioned design differs from simply encouraging high confidence, and directly optimizes the confidence-accuracy alignment measured by ECE.

\section{Experiments}

\subsection{Experimental Setup}

\textbf{Datasets.} We evaluate CARE on three widely adopted Medical VQA benchmarks:
\begin{itemize}[label=\textbullet,leftmargin=*]
    \item \textbf{VQA-RAD}~\cite{lau2018dataset}: Focused on radiology, comprising 315 clinician-annotated images and 3,515 query-answer pairs.
    \item \textbf{SLAKE}~\cite{liu2021slake}: A semantically-labeled, knowledge-enhanced dataset featuring 642 images and 14,000 bilingual pairs.
    \item \textbf{PathVQA}~\cite{he2020pathvqa30000questionsmedical}: Dedicated to pathology, containing 4,998 images and 32,799 question-answer pairs.
\end{itemize}

\noindent \textbf{Baselines.} We evaluate CARE against state-of-the-art reasoning-focused Medical MLLMs, categorized by parameter scale: 
(1) \textit{Compact-Scale Models ($\le$ 3B):} Including Med-R1-3B~\cite{lai2025medr1reinforcementlearninggeneralizable} and MedVLM-R1-2B~\cite{pan2025}.
(2) \textit{Standard-Scale Models (7B--8B):} Including Lingshu-7B~\cite{lasateam2025lingshugeneralistfoundationmodel}, MedVLThinker-7B~\cite{huang2025medvlthinkersimplebaselinesmultimodal}, Fleming-VL-8B~\cite{shu2025flemingvluniversalmedicalvisual}, and MedMO-8B~\cite{deria2026medmogroundingunderstandingmultimodal}.

\noindent \textbf{Evaluation Metrics.} We assess model performance along three dimensions:
\begin{itemize}[label=\textbullet,leftmargin=*]
    \item \textit{Diagnostic Accuracy:} We report Accuracy for closed-ended questions and Recall for open-ended questions, following standard Medical-VQA evaluation protocols.
    \item \textit{Confidence Calibration:} We adopt Expected Calibration Error (ECE)~\cite{guo2017calibration} to measure the alignment between model confidence and actual accuracy. Confidence scores are partitioned into $M$ equally spaced bins $B_m$, and ECE computes the weighted deviation:
    \begin{equation}
        \text{ECE} = \sum_{m=1}^{M} \frac{|B_m|}{N} |\text{acc}(B_m) - \text{conf}(B_m)|,
    \end{equation}
    where $N$ is the total number of samples. Lower ECE indicates better calibration.
    \item \textit{Hallucination Rate:} We employ Lingshu-32B~\cite{lasateam2025lingshugeneralistfoundationmodel} as a VLM judge to evaluate reasoning trajectories under a fixed prompt and rubric. The judge is run with deterministic decoding, and all methods are evaluated with identical inputs to ensure fairness. Each sample receives a normalized hallucination score $r_i$, and the overall HR is computed as:
    \begin{equation}
        \text{HR} = \frac{C}{\mathbb{E} \left[ \left(1 - r_i\right)^2 t_i \right]},
    \end{equation}
    where $t_i$ is the output token count and $C$ is a global normalization constant ensuring HR $\in [0,1]$. The squared term $(1-r_i)^2$ applies a quadratic penalty that disproportionately suppresses samples with high hallucination rates, amplifying the contribution of heavily hallucinated outputs to the overall score.
\end{itemize}

\subsection{Implementation Details}
CARE is built upon Qwen2.5-VL-7B-Instruct~\cite{bai2025qwen25vltechnicalreport}. The same model architecture serves as both the CoT data synthesizer in Section~\ref{sec:syn} and the policy model for subsequent training, ensuring consistency between the generated reasoning format and the optimization target. Training is conducted via full-parameter fine-tuning on $6\times$ NVIDIA A100 GPUs. During the \textit{SFT Cold Start}, we use the AdamW optimizer with a learning rate of $1 \times 10^{-5}$ and cosine annealing. In the \textit{GRPO RL Phase}, we generate $K=4$ rollouts per query with a batch size of 2 and \texttt{bfloat16} mixed precision. In the calibration reward of \textbf{CAR}, the penalty coefficient is set to $\lambda=0.5$.

\begin{table}[htbp]
\centering
\caption{Main results on Medical VQA benchmarks. The best and second-best results in each column are highlighted in \textbf{bold} and \underline{underlined}, respectively. The same applies to Table~\ref{tab:ablation} (\textbf{ACC} = accuracy, \textbf{ECE} = expected calibration error, \textbf{HR} = hallucination rate).}
\label{tab:main_results}
\resizebox{\textwidth}{!}{
\begin{tabular}{lccccccccc}
\toprule
\multirow{2}{*}{\textbf{Model}} & \multicolumn{3}{c}{\textbf{VQA-RAD}} & \multicolumn{3}{c}{\textbf{SLAKE}} & \multicolumn{3}{c}{\textbf{PathVQA}} \\
\cmidrule(lr){2-4} \cmidrule(lr){5-7} \cmidrule(lr){8-10}
& \textbf{ACC} $\uparrow$ & \textbf{ECE} $\downarrow$ & \textbf{HR} $\downarrow$ & \textbf{ACC} $\uparrow$ & \textbf{ECE} $\downarrow$ & \textbf{HR} $\downarrow$ & \textbf{ACC} $\uparrow$ & \textbf{ECE} $\downarrow$ & \textbf{HR} $\downarrow$ \\
\midrule
\multicolumn{10}{l}{\textit{Compact-Scale Models ($\le$ 3B)}} \\
\midrule
Med-R1-3B & 0.513 & 0.451 & 0.195 & 0.596 & 0.369 & 0.257 & 0.387 & 0.584 & 0.155 \\
MedVLM-R1-2B & 0.532 & 0.448 & 0.200 & 0.504 & 0.474 & 0.237 & 0.387 & 0.587 & 0.168 \\
\midrule
\multicolumn{10}{l}{\textit{Standard-Scale Models (7B--8B)}} \\
\midrule
MedMO-8B & 0.647 & \underline{0.264} & 0.365 & 0.816 & 0.277 & 0.368 & 0.563 & \underline{0.383} & 0.223 \\
Lingshu-7B & \underline{0.679} & 0.375 & 0.083 & \underline{0.831} & 0.322 & 0.105 & 0.619 & 0.522 & 0.098 \\
MedVLThinker-7B & 0.637 & 0.413 & \underline{0.071} & 0.678 & 0.424 & 0.099 & \underline{0.652} & 0.569 & 0.082 \\
Fleming-VL-8B & 0.668 & 0.333 & 0.073 & 0.819 & \underline{0.181} & \underline{0.089} & 0.629 & 0.455 & \underline{0.073} \\
\midrule
\textbf{CARE-7B (Ours)} & \textbf{0.767} & \textbf{0.202} & \textbf{0.048} & \textbf{0.873} & \textbf{0.115} & \textbf{0.070} & \textbf{0.689} & \textbf{0.290} & \textbf{0.059} \\
\bottomrule
\end{tabular}
}
\end{table}

\subsection{Experimental Results and In-depth Analysis}
\noindent \textbf{CARE simultaneously achieves the best accuracy, calibration, and lowest hallucination across all benchmarks.} As shown in Table~\ref{tab:main_results}, existing medical reasoning models consistently exhibit trade-offs among these three dimensions: strong accuracy often comes with poor calibration or high hallucination, and vice versa. No baseline ranks first across all three metrics on any single benchmark. \textbf{CARE} is the only model to do so consistently, achieving the highest accuracy (e.g., 0.873 on SLAKE, 0.767 on VQA-RAD) while simultaneously obtaining the lowest ECE and HR. The substantial ECE reduction (e.g., 0.115 on SLAKE, a 36\% relative improvement over the second-best Fleming-VL-8B) directly confirms the effectiveness of \textbf{CAR} in aligning confidence with accuracy. The consistently lowest HR (e.g., 0.048 on VQA-RAD) further validates our reverse-thinking data synthesis: answer-grounded reasoning paths reduce fabrication at the source by providing structured, goal-oriented CoT trajectories.

\begin{table}[tbp]
\centering
\caption{Ablation study results for different training stages.}
\label{tab:ablation}
\resizebox{\linewidth}{!}{
\begin{tabular}{lcccccccccccc}
\toprule
\multirow{2}{*}{\textbf{Training Stages}} & \multicolumn{4}{c}{\textbf{VQA-RAD}} & \multicolumn{4}{c}{\textbf{SLAKE}} & \multicolumn{4}{c}{\textbf{PathVQA}} \\
\cmidrule(lr){2-5} \cmidrule(lr){6-9} \cmidrule(lr){10-13}
& \textbf{Open $\uparrow$}& \textbf{ECE $\downarrow$}& \textbf{Closed $\uparrow$}& \textbf{ECE $\downarrow$}& \textbf{Open $\uparrow$}& \textbf{ECE $\downarrow$}& \textbf{Closed $\uparrow$}& \textbf{ECE $\downarrow$}& \textbf{Open $\uparrow$}& \textbf{ECE $\downarrow$}& \textbf{Closed $\uparrow$}& \textbf{ECE $\downarrow$}\\
\midrule
Training-Free & 0.483 & 0.485 & 0.658 & 0.341 & 0.513 & 0.456 & 0.690 & 0.287 & 0.150 & 0.823 & 0.664 & 0.326 \\
SFT           & 0.520 & 0.468 & 0.629 & 0.359 & 0.803 & 0.188 & 0.767 & 0.225 & \underline{0.349} & \underline{0.640} & 0.857 & 0.133 \\
RL            & \underline{0.563} & \underline{0.407} & \textbf{0.864} & \textbf{0.096} & \underline{0.819} & \underline{0.176} & \textbf{0.861} & \textbf{0.119} & 0.166 & 0.727 & \textbf{0.955} & \textbf{0.020} \\
SFT+RL        & \textbf{0.620} & \textbf{0.363} & \underline{0.658} & \underline{0.323} & \textbf{0.881} & \textbf{0.112} & \underline{0.779} & \underline{0.209} & \textbf{0.421} & \textbf{0.561} & \underline{0.863} & \underline{0.121} \\
\bottomrule
\end{tabular}
}
\end{table}

\noindent \textbf{Different question types favor different training configurations.} Table~\ref{tab:ablation} reveals a clear pattern: for closed-ended questions, RL alone achieves the best accuracy and ECE across all three benchmarks (e.g., 0.864 ACC and 0.096 ECE on VQA-RAD), as the constrained answer space allows RL exploration to converge efficiently without SFT initialization. For open-ended questions, however, SFT+RL consistently dominates (e.g., 0.881 on SLAKE Open, 0.421 on PathVQA Open), since generating free-form diagnostic reasoning requires the structured CoT foundation established during SFT. Notably, SFT alone already yields substantial open-ended gains (e.g., SLAKE Open rises from 0.513 to 0.803), confirming that the CoT cold-start is the primary driver for open-ended performance, while RL further refines calibration on top of it. The results also reflect the complementary roles of the two stages: Medical-CoT provides a stable reasoning and answer-extraction format, especially for open-ended questions, while the RL stage further optimizes answer correctness and calibration through CAR.

\begin{figure}[tb]
\includegraphics[width=\textwidth]{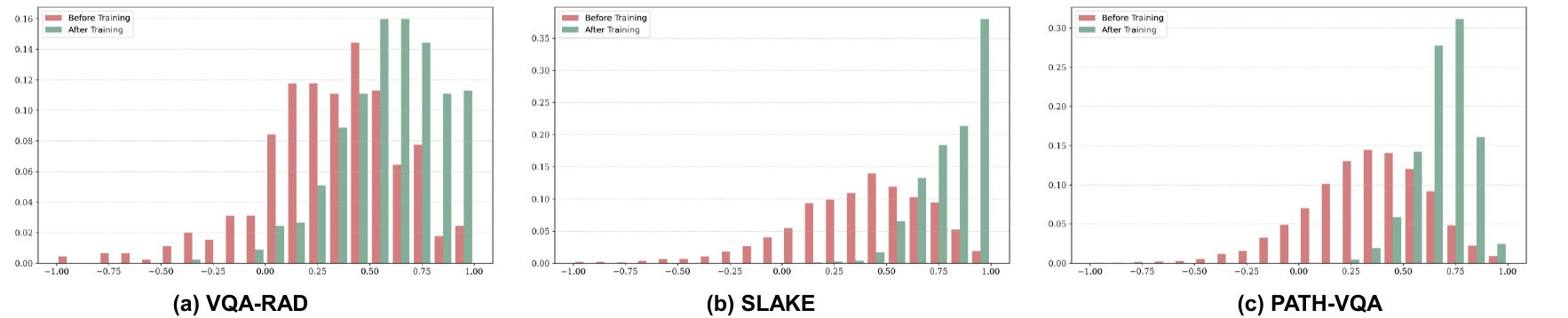}
\caption{\textit{Confidence distributional shift.} Red and green bars represent the normalized confidence scores pre- and post-training, respectively. After training, the mean shifts rightward with increased magnitude, reflecting improved confidence-accuracy alignment.} \label{fig3}
\end{figure}

\noindent \textbf{CAR improves confidence-accuracy alignment rather than merely increasing confidence.}
Figure~\ref{fig3} compares the normalized confidence distributions before and after \textbf{CAR} alignment. The post-training distribution shifts toward higher-confidence regions, but this shift should be interpreted together with the simultaneous ECE reductions in Table~\ref{tab:main_results}. Since ECE directly measures the gap between confidence and empirical accuracy, the improved ECE indicates that the confidence shift is better aligned with diagnostic correctness rather than being a simple increase in likelihood. This supports the design of \textbf{CAR}, which rewards confident correct predictions while penalizing overconfident incorrect ones.

\section{Conclusion}
We present \textbf{CARE}, a framework that addresses confidence miscalibration in RFT-based medical reasoning models. By combining a scalable Medical-CoT synthesis pipeline with a Confidence-Aware Reward (\textbf{CAR}) mechanism within GRPO, \textbf{CARE} explicitly aligns the model's expressed confidence with diagnostic accuracy during RL optimization. Experiments across three Medical VQA benchmarks show that \textbf{CARE} achieves the best diagnostic accuracy while simultaneously obtaining the lowest ECE and Hallucination Rate, demonstrating that accuracy and calibration can be jointly improved rather than traded off. These results suggest that incorporating confidence signals into the reward design is an effective path toward reliable medical AI systems. We sincerely hope this work encourages further exploration of calibration-aware training objectives in safety-critical domains beyond medical VQA.

\begin{credits}
\subsubsection{\ackname} This work was supported by the National Natural Science Foundation of China under Grant 42394060 and 42394064, Ant Group Research Fund, and the Zhejiang University - Jolly Pharmaceutical Joint R\&D Center for Intelligent Empowerment in Food and Medicine.

\subsubsection{\discintname}
The authors have no competing interests to declare.
\end{credits}
%
%
%
%
\bibliographystyle{splncs04}
\bibliography{Paper-6558}
\end{document}